# Using a Machine Learning Approach for Computational Substructure in Real-Time Hybrid Simulation

Elif Ecem Bas[1], Mohamed A. Moustafa[1], David Feil-Seifer[2], and Janelle Blankerburg[2]

[1] Department of Civil and Environmental Engineering
College of Engineering

[2] Department of Computer Science and Engineering
College of Engineering

University of Nevada, Reno, 1664 N. Virginia Street Reno, NV 89557

**ABSTRACT**
Hybrid simulation (HS) is a widely used structural testing method that combines a computational substructure with a numerical model for well-understood components and an experimental substructure for other parts of the structure that are physically tested. One challenge for fast HS or real-time HS (RTHS) is associated with the analytical substructures of relatively complex structures, which could have large number of degrees of freedoms (DOFs), for instance. These large DOFs computations could be hard to perform in real-time, even with the all current hardware capacities. In this study, a metamodeling technique is proposed to represent the structural dynamic behavior of the analytical substructure. A preliminary study is conducted where a one-bay one-story concentrically braced frame (CBF) is tested under earthquake loading by using a compact HS setup at the University of Nevada, Reno. The experimental setup allows for using a small-scale brace as the experimental substructure combined with a steel frame at the prototype full-scale for the analytical substructure. Two different machine learning algorithms are evaluated to provide a valid and useful metamodeling solution for analytical substructure. The metamodels are trained with the available data that is obtained from the pure analytical solution of the prototype steel frame. The two algorithms used for developing the metamodels are: (1) linear regression (LR) model, and (2) basic recurrent neural network (RNN). The metamodels are first validated against the pure analytical response of the structure. Next, RTHS experiments are conducted by using metamodels. RTHS test results using both LR and RNN models are evaluated, and the advantages and disadvantages of these models are discussed.

## INTRODUCTION

Hybrid simulation (HS) is developed to answer the need for realistic dynamic testing of structures that combines experimental and analytical models and benefits from their advantages simultaneously. HS was first introduced by Takanashi et al. [1], where it was defined as "on-line testing" and the structural system modeled as a discrete spring-mass model within the time domain. Since then, many researchers were studied in different areas to enlarge the HS capabilities such as developments in numerical integration methods [2–6], substructuring techniques [7–9], delay compensation, and error mitigation [10–12]. With these developments, this experimental technique became more reliable, accurate, efficient, and cost-effective for large-scale and full-scale real-time dynamic testing [13].

During the dynamic analysis in HS/RTHS, the equation of motion for the coupled experimental-computational simulations is usually solved by direct numerical integration algorithms. However, there are still some limitations that exist for these integration algorithms for complex structures where larger degrees of freedoms are involved along with numerical and/or experimental nonlinearities. These limitations could affect the performance and reliability of the HS/RTHS. To avoid these difficulties and improve the performance of the HS/RTHS, machine learning algorithms could be an alternative way to represent the analytical substructure. Machine learning is the science of programming computers so that they can learn from data [14]. The use of machine learning algorithms or metamodels became popular recently in engineering problems, with the growing complexity of finite element (FE) models [15]. Even with the computational powers that we have today, the computational cost for FE models, which have large numerical systems, could be very high, and this led researchers to develop alternative representations of these FE models to predict the dynamic response of the simulation.

The main idea of this paper is to represent an entire concentric braced frame (CBF) computational substructure, i.e. columns, beam, mass, and damping, with a metamodels that is developed using a machine learning algorithm to define the analytical substructure response in HS. Moreover, the inherent servo-hydraulic dynamics in the HS system could lead to a time delay in response to the command of displacement, which generates inaccurate results, especially in RTHS [16]. This time delay is usually eliminated by using a proper delay compensator. In this paper, time delay with the actuator input and the feedback are eliminated within the metamodel instead of using a time delay compensator. To do these, a dataset for the training model is obtained by the pure analytical model time history analysis of a one-bay one-story CBF, under earthquake excitation. The model is considered as batch learning since all the dataset are provided offline. A linear regression (LR) model and a recurrent neural network (RNN) model are considered to develop the metamodels, which predicts the input displacement value for the actuator. Once the metamodels are trained, these are first compared with the pure analytical FE model response. The FE model response is considered to be the exact solution of the system. Then, the metamodels are incorporated into the HS loop to conduct preliminary linear tests, where the HS test results and model responses can be compared against the exact values from pure FE analysis as well.

**SYSTEM COMPONENTS AND CAPABILITIES**

A compact HS setup is designed and constructed in the Large-Scale Structures Laboratory (LSSL) at the University of Nevada, Reno. This small-scale setup is developed for CBF demonstrations, educational purposes, and tackle new research problems pertaining to computational challenges for HS/RTHS. The analytical substructure can be modeled using either Simulink or OpenSees [17] platform by using proper FE techniques. Moreover, the HS system is capable of running both real-time and pseudo-dynamic (slow) experiments. More details about the HS system development and verification can be found in [18].

The system consists of: (1) load frame with a dynamic actuator run by an isolated hydraulic pump; (2) MTS STS controller with 4-channels with 2048 Hz clock speed; (3) real-time high performance Simulink machine (Speedgoat xPC Target); (4) Windows machine (Host PC) for MATLAB, OpenSees, and the HS middleware OpenFresco [19]; and (5) SCRAMNet ring that provides shared memory locations for real-time communication.

The xPC Target provides a high-performance host-target prototyping environment that enables the researcher to connect the Simulink and Stateflow models to physical systems. It sends and receives data from the controller. The controller (STS 493 Hardware Controller) has 4-channels with 2048 Hz clock speed and controls the motion of the actuator. Currently, it has only one channel connected since it is only controlling one actuator, but it is capable of controlling 4 actuators. STS Host PC is where the basic controller properties are controlled with the graphical user interface of the MTS 493 controller [20].

The load frame is the experimental setup for the HS system. The dynamic actuator of the system has 7 kips (31.14 kN) maximum load capacity, which has (±1 in.) stroke. The peak velocity of the actuator at no load is 338.84 mm/sec (13.34 in/sec). The isolated hydraulic power supply system has 8.71 lt/min (2.3 gpm) pumps, and the reservoir capacity of oil volume is 56.78 lt (15 gallons).

**MODELING ASSUMPTIONS**

The one-story one-bay steel braced frame is selected for both verifications and evaluations of the system challenges. CBFs are ideal for HS testing since the columns and beams are not expected to be damaged during an earthquake and can be accurately modeled in the computer along with the mass and the damping forces (i.e. analytical substructure), while the braces can be physically tested to capture buckling accurately and low-cycle fatigue induced rupture (i.e., experimental substructure). Figure 1 illustrates the CBF substructuring for HS testing. The experimental setup allows for using a small-scale brace as the physical substructure along with a steel frame at the prototype full-scale for the analytical substructure. The

HS of this model has been verified against the pure analytical solution of frame, and these results are also included in this section.

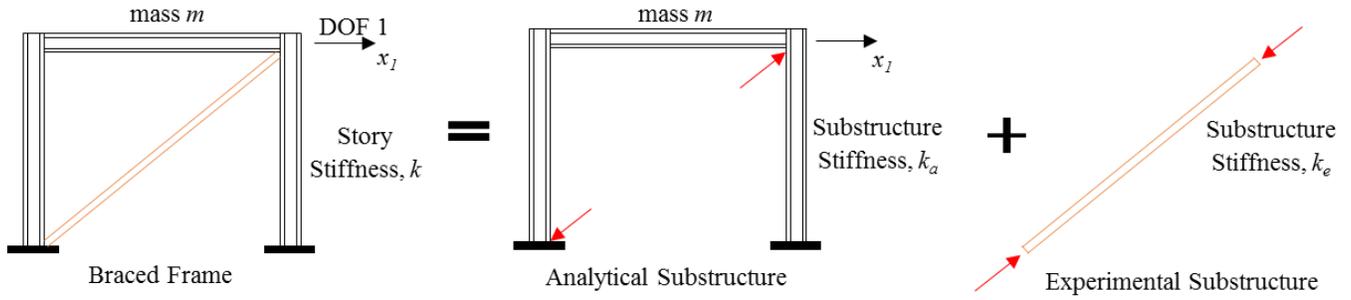

Figure 1. Model and substructuring of CBFs for HS testing.

Two machine learning algorithms are pursued in this study to develop metamodels and explore possibilities to compile a successful HS test. First, the linear regression (LR) method is used to train a model. Secondly, a recurrent neural network (RNN) with different hidden layers and time step delays is modeled and tested for performance. Both algorithms are compared with the pure analytical models first. Then, following the assessment of their performance in the pure analytical model predictions, the metamodels are further tested in the HS loop with real feedback from the actuator that is free to move without specimen attached as explained later.

In general, an LR model predicts by simply computing a weighted sum of the input features, plus a constant called bias term [14], shown in Equation (1). Here $\hat{y}$ is the predicted value, $n$ is the number of features. $x_i$ is the i$^{th}$ feature value, and $\theta_j$ is the j$^{th}$ model parameter (including the bias term $\theta_0$ and the feature weights term).

$$\hat{y} = \theta_0 + \theta_1 x_1 + \theta_2 x_2 + \cdots + \theta_n x_n \tag{1}$$

The LR model is trained by using a pure analytical solution of the CBF. For the training, five features are selected to be the input of this metamodel to predict the output as a command displacement of the experimental substructure of the HS system. The features for training are selected as: ground motion acceleration, displacement feedback value of the brace, force feedback value of the brace, one step behind the value of the predicted displacement, and two steps behind the value of the predicted displacement. First, the pure analytical solution is used as training data, generated by using the pure analytical solution responses and assuming that there is a 28-time step time delay for the feedback values. Moreover, since it is hard to explore the exact delay from the actuator feedback, this model is run first in the HS loop without including the displacement feedback from the actuator. However, the displacement feedback of the actuator is recorded to be used in the next training phase. Then, a more refined model is generated by using the "real" displacement and force feedback, and the other three features that are used for the training phase.

For the RNN model, three different models are trained and generated. An RNN is an artificial neural network that allows exhibiting temporal dynamic behavior since the model uses its memory to process the sequence of inputs [21]. For the RNN models, the inputs are ground motion acceleration and force feedback. Basic RNN models are generated by using different hidden layers to find the optimum hidden layer number for an accurate response.

Once the models are trained, each method is evaluated, and the results are shown. Root mean square error (RMSE) is calculated to evaluate the performance of the model predictions. The first verification is done for a pure analytical machine learning algorithm to see the model capabilities offline. Here, the prediction is made for each step while running the simulation. Once all the models are compared for the pure analytical solution, the one with the least RMSE error will be put in the HS loop. The first validation for the HS system is where the metamodel is compiled in the xPC and is tried with "fake" feedback with actuators on the system (online) to see the actuator performance. After this, the "real" feedback is fed into the metamodel, and the system response is evaluated against the pure analytical system response. The RMSE values are calculated for these cases as well as discussed next.

## MODEL PARAMETERS FOR HS

The analytical substructure for the HS model consists of two columns (W14x311) with fixed end conditions and a beam (W36x150), which also has moment connections to the columns. Both columns and the beam are considered linear elastic elements, wherein braced frames the nonlinearity is commonly limited only to the braces. The mass and the damping of the system are considered to be part of the analytical substructure. The one-bay one-story braced frame is simplified as single degree of freedom (SDOF) for validation and modeling purposes. The dynamic properties of the SDOF model is as defined where M is the mass matrix, and K is the stiffness matrix given by Equations (2) and (3).

$$M = [m] \qquad (2)$$

$$K = [k_a + k_e] \qquad (3)$$

For this study, the story mass $m$ is selected to be 1.75 kN-sec2/mm, and the frame stiffness ($k_a$) is determined from the frame sections to be 176.75 kN/mm. The axial brace stiffness, along its local axis, is 1224.1 kN/mm. The resulting natural period of the frame is 0.294 sec. The structure is assumed to have an inherent damping of 2% modeled using Rayleigh damping. The equation of motion of the full system can be written as shown in Equation (4).

$$m\ddot{x} + c\dot{x} + kx = -m\ddot{u}_g \qquad (4)$$

$c$ is the inherent viscous damping of the structure; $x(t)$, $\dot{x}(t)$, and $\ddot{x}(t)$ are the displacements, velocity, and acceleration response, respectively, and $\ddot{u}_g$ is the ground motion acceleration. For HS, the equation is rewritten such that $k_a$ is the frame stiffness representing the analytical substructure, and $k_e$ is the experimental stiffness which represents the brace stiffness. The brace stiffness is transformed into global coordinates. In the HS case, the feedback from the experimental substructure replaces the $k_e x_i$ term in Equation (5).

$$m\ddot{x}_i + c\dot{x}_i + (k_a + k_e)x_i = -m\ddot{u}_g \qquad (5)$$

The pure analytical model that is developed in Simulink is using the Chang integration algorithm and uses a 1/2048 second time step [22]. The time step of the integration is selected to be the same as the controller time step to synchronize the data transfer. For the purpose of dynamic analysis, and later the HS seismic testing, a typical California ground motion is selected and used, which is the 1940 El-Centro ground motion excitation shown in Figure 2a. From pure dynamic analysis, the brace displacement and brace force histories are obtained and shown in Figures 2b and 2c, respectively. It is noted that such time history data is modified and used as the training data, to represent the feedback data, for the machine learning approach. To do so, a 28-time step delay is considered along with the pure analytical model data when used for the training and validation phases of pure analysis of the metamodel. As previously mentioned, metamodel is updated with the "real" feedback data that is obtained from free moving actuator.

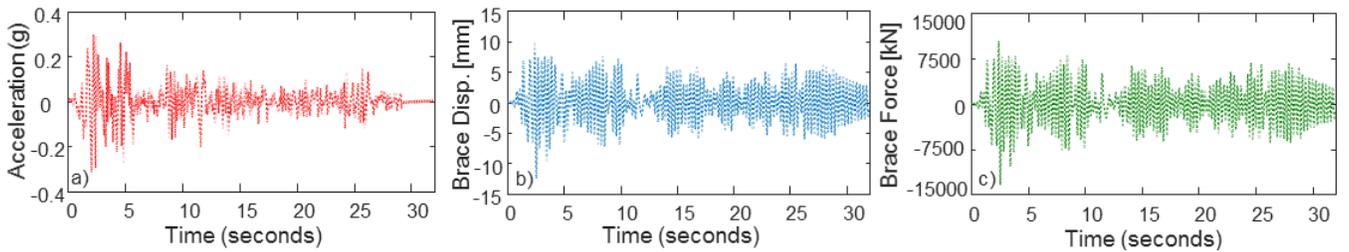

Figure 2 (a) El-Centro ground motion acceleration; and brace results from pure analysis: (b) displacement history, and (c) force history

## VALIDATION FOR RTHS WITH FE MODEL

In this section, before getting into the HS test with metamodel, a brief summary of the validation tests that are conducted to validate the SCRAMNetGT card connection between the computers and the controller are presented. These tests were conducted in real-time, i.e. RTHS. For the offline validation test, the feedback from the experimental model was considered as it is coming from the command of the system. The displacement command was multiplied with the constant stiffness value of the specimen, and these validation tests were considered for linear elastic experimental material. Figure 3a shows the brace displacement time history comparison between the pure analytical model and offline RTHS response. In this case the RTHS response has an RMSE value as 5.4%.

Moreover, the system capabilities were also verified, where an actual feedback from the actuator is fed into the analytical model. In this validation phase, instead of using a linear specimen, the feedback displacement from the free-to-move actuator is multiplied with a constant value equals to the stiffness of the brace, i.e. experimental substructure, in order to get the force feedback for the HS experiment. This case also represents a hypothetical linear-elastic brace material since the stiffness was used as a constant value. The actuator delay was corrected and compensated for using the Adaptive Time Series (ATS) compensator [10]. Figure 3b shows the displacement history comparisons for the online validation case of RTHS. When the actuator is online, and the real feedback is considered in the HS system (as it should be), the RMSE value is calculated to be 9.4%.

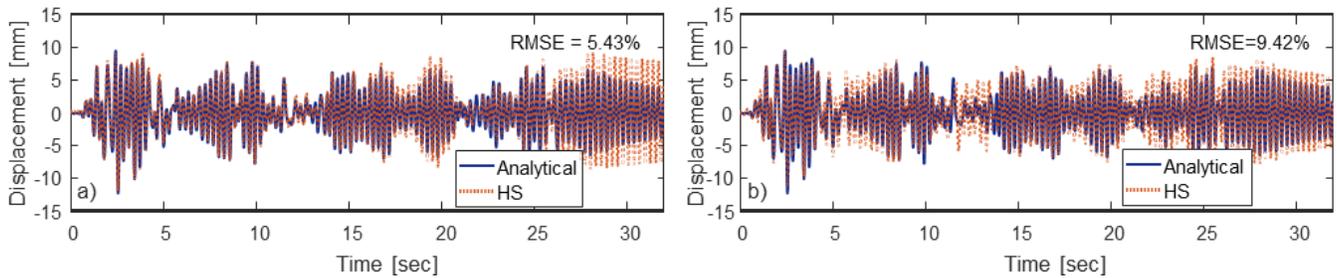

Figure 3 HS validation test results: (a) for the offline case with FE model, (b) for the online case with FE model without specimen

## METHODOLOGY FOR LINEAR REGRESSION ALGORITHM

Firstly, the LR method is used in order to train a machine learning metamodel. The input features of this metamodel are selected to be: earthquake excitation, displacement feedback of the brace, force feedback of the force, and one- and two-previous steps of the history of predicted displacement value of the brace. Once the model is trained, the model parameters are obtained, and a Simulink model with Matlab Function that contains the LR model is prepared as illustrated in Figure 4. This model is compared with the pure analytical model solution that is presented above, i.e. FE model.

$$x_{prediction,i+1} = f(\ddot{x}_g, x_{feedback,i}, F_{feedback,i}, x_{prediction,i}, x_{prediction,i-1}) \qquad (6)$$
$$= \theta_1 \ddot{x}_g + \theta_2 x_{feedback,i} + \theta_3 F_{feedback,i} + \theta_4 x_{prediction,i} + \theta_5 x_{prediction,i-1}$$

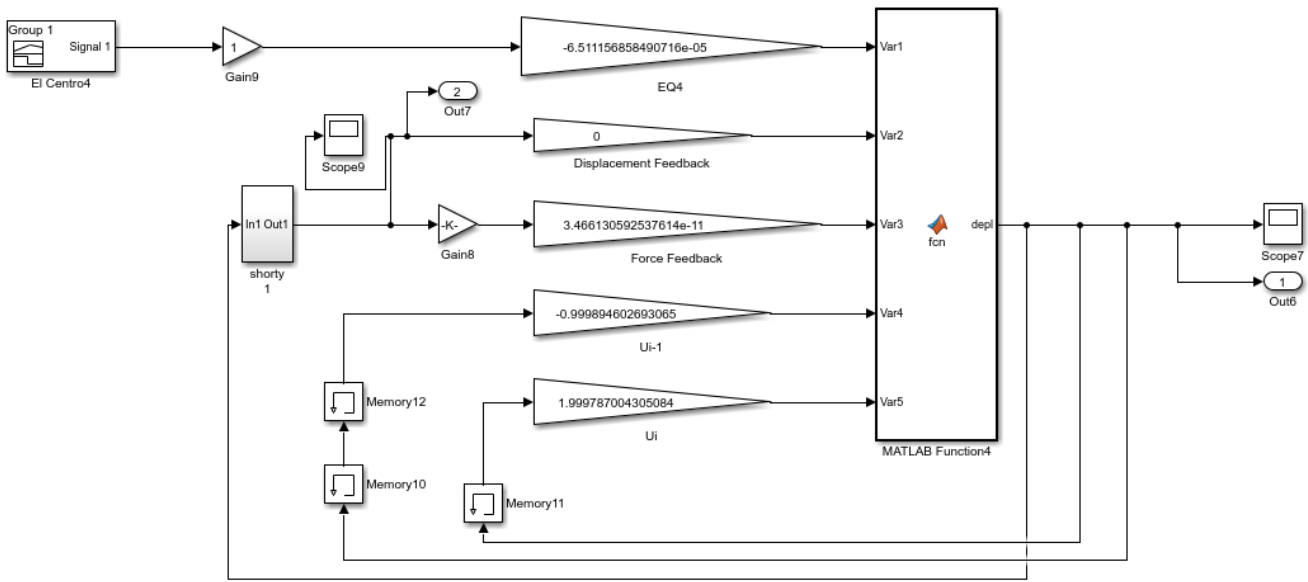

Figure 4 Machine Learning Model (Pure Analytical Model)

In Figure 5, the brace displacement response of the FE model and metamodel with LR are compared when both models are used in a pure analytical setting, i.e. not from a HS test. The FE model response is considered as the exact solution of the system. The RMSE value for this comparison using the LR model is 0.15%, which indicates that the LR model performs well for the pure analytical case. It is noted that this model is trained with "real" displacement feedback that is obtained from the HS setup as previously discussed. However, in this pure analytical metamodel, displacement feedbacks are generated by using a 28-time step delay from the calculated (predicted) displacement.

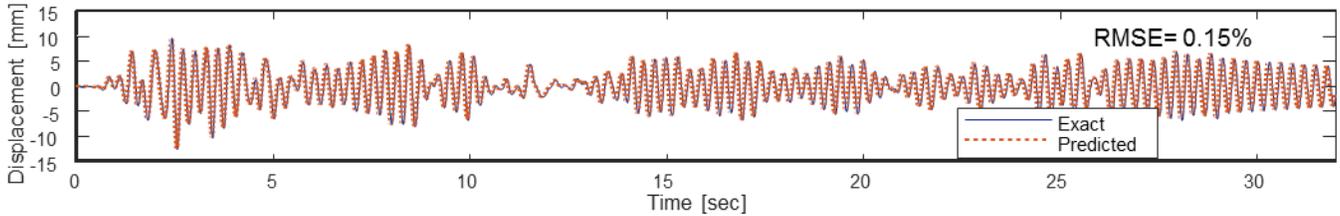

Figure 5 Comparison of results from pure analysis using FE model versus LR metamodel

Once the pure analytical model is verified for the metamodel, the metamodel is compiled in the xPC Target so that it can be tested on a different hardware. For this model, again the calculated (predicted) value is modified with a 28-time step delay and fed back into the model, which means there is no feedback from the actuator. However, the hydraulics of the HS system are on and the actuator is free to move with the given input command, which is the predicted displacement of the brace. Figure 6 shows the brace displacement response as obtained from RHTS offline tests incorporating the LR metamodel, which is compared with the pure analytical FE results. The results are comparable as before which further validates the LR model.

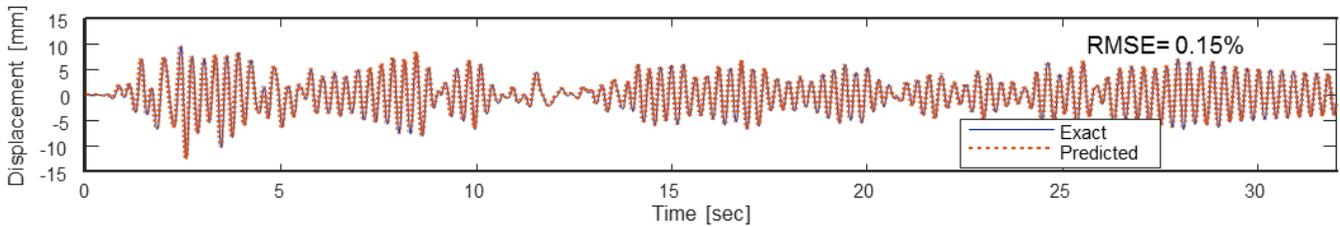

Figure 6 Comparison of brace displacement from offline HS validation test with the LR model (w/o actuator feedback) and pure analysis

It can be seen from the above verifications that the metamodel is giving adequate predictions when compared with the exact solution. Next, the actuator behavior is also investigated that no unexpected behavior is observed during actual tests. The metamodel is required to be validated with the real feedback from the actuator. Thus, no specimen is used here but the actuator displacement feedback is multiplied with a constant stiffness value of the brace to produce an equivalent force feedback of a linear elastic specimen. The Simulink configuration that is used for this validation phase is shown in Figure 7. This model is also compared with the pure analytical FE model solution, which can be seen in Figure 8. The comparison shown in the figure corresponds to a RMSE value of 0.066%, which concludes that the RTHS test with the LR model is acceptable and the performance is very comparable to pure analytical cases.

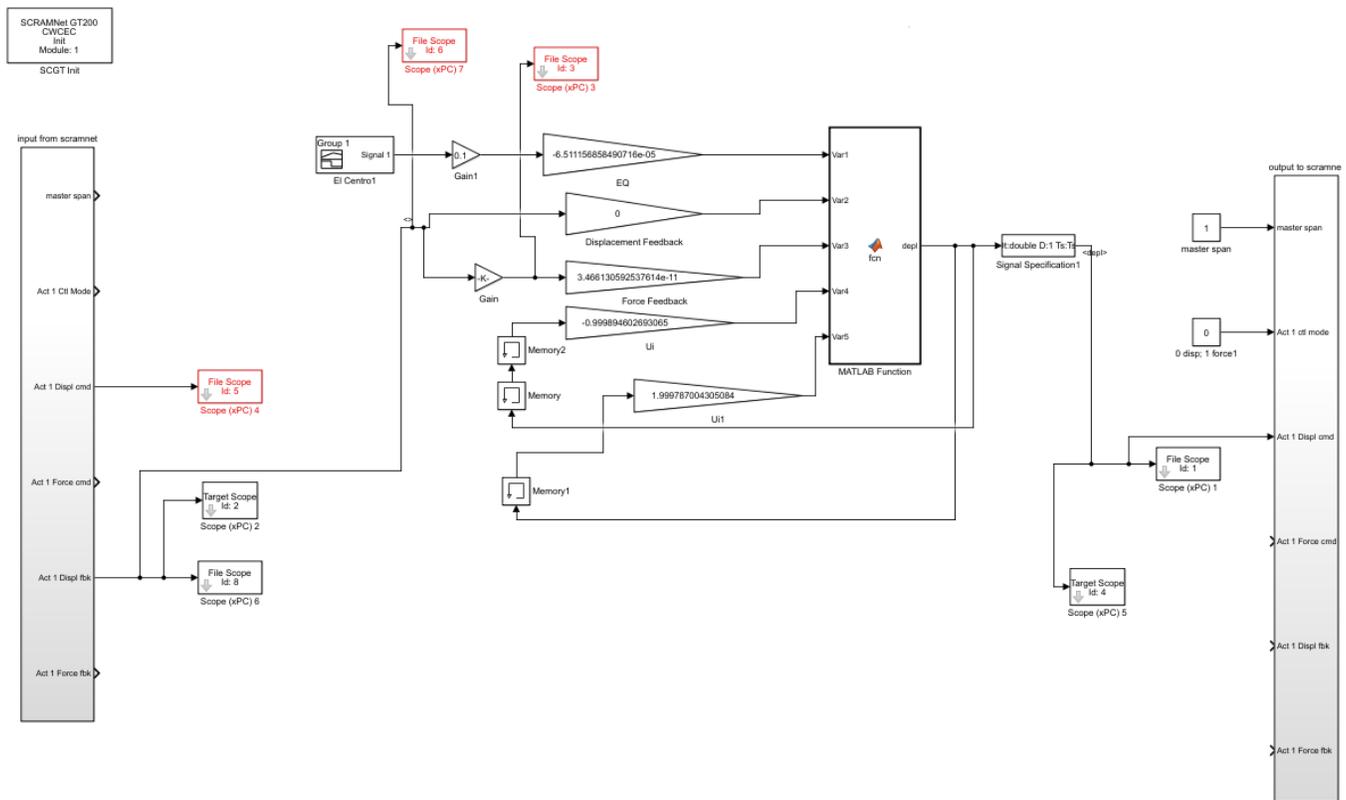

Figure 7 Simulink configuration of HS with the LR model

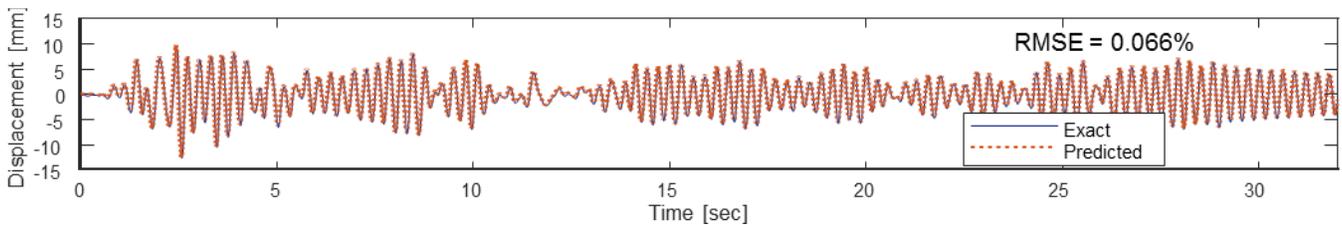

Figure 8 Comparison of brace displacement from online HS validation test with the LR model (w/ actuator feedback) and pure analysis

**METHODOLOGY FOR RECURRENT NEURAL NETWORK ALGORITHM**
As a second metamodeling algorithm, an RNN is used since this problem is a time series prediction. RNN is a class of nets that can predict the future, which can analyze time series [14]. The RNN model representation is shown in Figure 9, which shows two inputs and one output. In this paper, the two input features are defined as: the earthquake excitation ($x_1$) and the force feedback of the brace ($x_2$) where the output for a HS test setting is the command signal for the brace displacement (y). For this method, since it has a connection pointing backward by default, there is no need to define an input for either one- or/and two-step behind predicted displacement value of the brace.

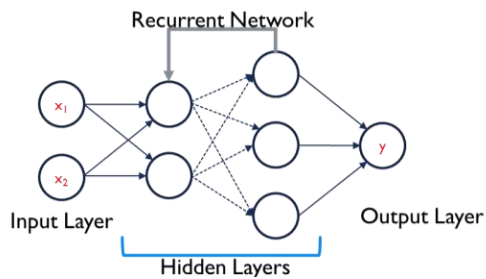

Figure 9 RNN Architecture

The methodology in an RNN is defined as follows. Each recurrent neuron has two sets of weights where one for the inputs $\mathbf{x}_{(t)}$ and the other for the outputs of the previous time step, $\mathbf{y}_{(t-1)}$, and these weight vectors are $\mathbf{w}_x$ and $\mathbf{w}_y$, respectively. The output of a single recurrent neuron for a single instance can be defined as shown in Equation (7) [14]. In the following equation $b$ is the bias term and $\emptyset(.)$ is the activation function.

$$\mathbf{y}_{(t)} = \emptyset\left(\mathbf{x}_{(t)}^T \cdot \mathbf{w}_x + \mathbf{y}_{(t-1)}^T \cdot \mathbf{w}_y + b\right) \tag{7}$$

Basic RNN is used in order to investigate the method. Three models with different number of hidden layers are generated to evaluate the performance. The model properties and names are summarized in Table 1. Five hidden layers were selected to start with and increased to 10 and 20 at last. These models are compared with the pure analytical FE model response, and the RMSE values are obtained as discussed next.

*Table 1 RNN Model Parameters*

| Model Name | Hidden Layers |
|---|---|
| Model-1 | 5 |
| Model-2 | 10 |
| Model-3 | 20 |

In Figure 10, the RNN model response that is trained with 5 hidden layers is shown. It can be seen from the responses that the RNN model cannot catch the peak displacement responses at the beginning. Moreover, a zoomed-in view of the response show that there are serious overfitting problems for some data points. On the other hand, that metamodel can capture the dynamics adequately. The RMSE value for this model is 4.23%, which can be considered to be overall adequate.

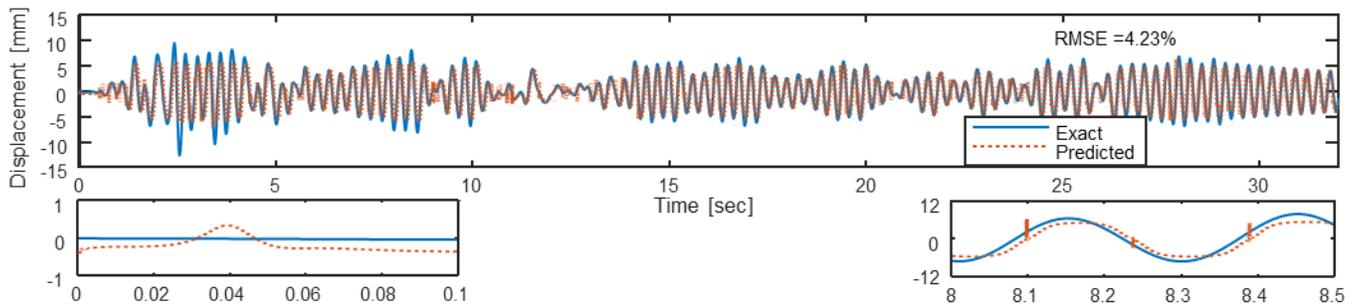

Figure 10 Response comparison of FE model and pure analytical RNN model with 5 hidden layers (Model 1)

After assessing the performance of Model 1, the hidden layer number is increased to 10 and results from Model 2 are shown in Figure 11. As it can be seen in Figure 11, the model performance is enhanced considerably when compared to Model 1. The dynamics of the structure are captured well, and the RNN model catches the peak displacements. The RMSE value is decreased to 3.74%. However, when the initial points are investigated, there is a severe oscillation in the behavior that can be observed as shown in Figure 11. Moreover, along the displacement history, there is still overfitting in some data points.

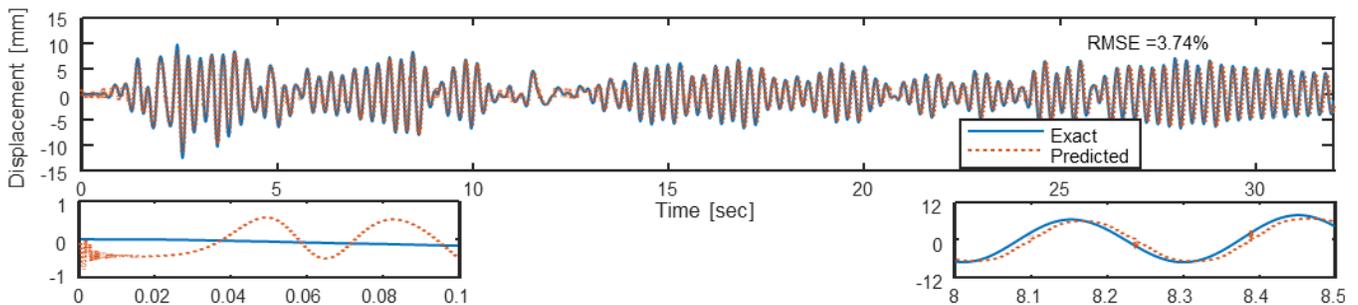

Figure 11 Response comparison of FE model and pure analytical RNN model with 10 hidden layers (Model 2)

Lastly, an RNN model with 20 hidden layers is trained, i.e. Model 3. The brace displacement response comparison for Model 3 against FE model is shown in Figure 12. The RMSE value has not been improved as compared to Model 2 and slightly increased to 3.88%. The performance also is not improved where this model still do not capture the peaks accurately. Moreover, the noise level at the beginning of the movement is also increased. The RNN models are shown here as potential alternative metamodels but given that RNN models performance is not significantly better than LR, no HS tests are conducted here using the RNN and only the analytical results above are provided for completeness.

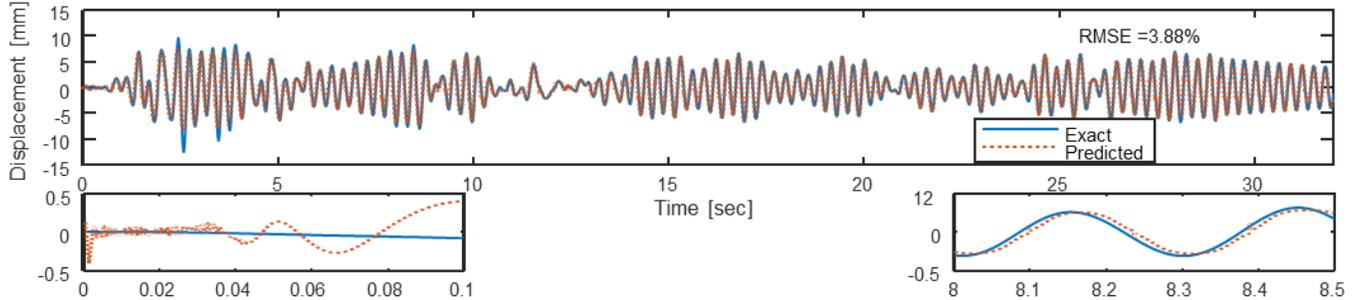

Figure 12 Response comparison of FE model and pure analytical RNN model with 20 hidden layers (Model 3)

## SUMMARY AND CONCLUSIONS

In this paper, machine learning algorithms are used for first time to model structural response of potential HS analytical substructure as a new alternative way to replace finite element models. The idea is to represent all analytical substructure components, i.e. frame columns, beam, mass, and damping in this case, with a metamodel, to reduce the computational time spent on the analytical substructure. Two different machine learning algorithms are used, i.e. LR and a basic RNN. Besides the computational time benefits, using metamodels for HS testing could eliminate the delay between the displacement command and feedback in the actuator within the model without using any time delay compensator.

The HS setup and the validation of the setup using a FE model is first presented. The dataset to train the machine learning algorithms are obtained from a one-bay one-story braced frame response under the El-Centro earthquake. The columns and brace are defined as an analytical substructure where the brace is an experimental substructure. In this study, both analytical and experimental substructures are defined to be linear elastic.

Firstly, an LR model is generated. For this model, five inputs features are used which are: earthquake excitation, displacement feedback of the brace, force feedback of the brace, and one- and two-previous steps from history of predicted displacement value of the brace. The FE model responses from pure analysis are used as dataset. Force and displacement feedbacks are generated by including a 28-time step delay to the analytical model response. However, it was observed that measuring the "online" actuator delay and the amplitude error is not possible. Thus, a preliminary metamodel should be generated initially and used in a HS setting in order to get "real" feedback. Once the "real" feedback is obtained, the final metamodel is generated to use for more representative HS tests. The metamodel with the LR algorithm is verified with a pure analytical FE model response. When the RMSE value of the predicted response history is found to be satisfactory, the model was tested within the HS loop but without including the "real" feedback data in the machine learning algorithm to check the actuator response offline. Once this gives accurate results, the "real" feedback is also included in the machine learning algorithm. The LR model has performed well and provided adequate and comparable results as pure analytical response.

Secondly, RNN is also trained to represent the analytical substructure in the HS system. Three different models are trained where the hidden layers are changed to be 5, 10, and 20, respectively. The RNN model with 10 hidden layers, has given the most accurate result, which has the least RMSE value when compared to pure FE analytical response. The RNN models are able to capture the dynamics of the structure accurately. However, when the response is investigated in more detail, it can be seen that there are some overfitting on some data points. In addition, lots of oscillations were observed in the initial part of the response. Since these errors can cause critical problems for the actuator, the RNN model was not considered further to use in the HS loop but yet presented here to outline its limitations.

In conclusion, for a linear-elastic brace and linear-elastic analytical substructure, the new concept of HS testing using a machine learning algorithm is validated, for the first time, using actual HS tests. The analytical substructure represented with an LR algorithm gives satisfactory results which is very promising as compared to FE models. On the other hand, RNN models are not yet validated to use within the HS loop and more work is needed on RNN to reduce the noise and overfitting issues. For future work, a crucial step is to generalize the machine learning concept for HS for models with nonlinear responses. Moreover, different machine learning algorithms can also be investigated in the search for more effective and accurate methods for HS testing.

**REFERENCES**
[1] Takanashi, K., Udagawa, K., Seki, M., Okada, T., Hisashi, T.: Non-Linear Earthquake Response Analysis of Structures By a Computer-Actuator On-Line System. Bull. Earthq. Resist. Struct. Res. Cent. (No.8). Japan Inst. Ind. Sci. Univ. Tokyo. (1975)
[2] Ahmadizadeh, M., Mosqueda, G.: Hybrid simulation with improved operator-splitting integration using experimental tangent stiffness matrix estimation. J. Struct. Eng. 134, 1829–1838 (2008). doi:10.1061/(ASCE)0733-9445(2008)134:12(1829)
[3] Jinting, W., Liqiao, L., Fei, Z.: Efficiency analysis of numerical integrations for finite element substructure in real-time hybrid simulation. 17, 73–86 (2018). doi:https://doi.org/10.1007/s11803-018-0426-0
[4] Kolay, C.: Parametrically Dissipative Explicit Direct Integration Algorithms for Computational and Experimental Structural Dynamics, (2016)
[5] Bonnet, P.., Williams, M.S., Blakeborough, A.: Evaluation of numerical time-integration schemes for RTHS. 1467–1490 (2014). doi:10.1002/eqe
[6] Chen, C., Ricles, J.M.: Development of Direct Integration Algorithms for Structural Dynamics Using Discrete Control Theory. J. Eng. Mech. 134, 676–683 (2008). doi:10.1061/(ASCE)0733-9399(2008)134:8(676)
[7] Shao, X., Mueller, A., Mohammed, B.A.: Real-Time Hybrid Simulation with Online Model Updating: Methodology and Implementation. J. Eng. Mech. 142, 1–19 (2015). doi:10.1061/(ASCE)EM.1943-7889.0000987.
[8] Mettupalayam, S., Reinhorn, A.: Real Time Dynamic Hybrid Testing Using Force-Based Substructuring. (2006)
[9] Zhou, M.X., Wang, J.T., Jin, F., Gui, Y., Zhu, F.: Real-time dynamic hybrid testing coupling finite element and shaking table. J. Earthq. Eng. 18, 637–653 (2014). doi:10.1080/13632469.2014.897276
[10] Chae, Y., Kazemibidokhti, K., Ricles, J.M.: Adaptive time series compansator for delay compensation of servo-hydraulic actuator systems for real-time hybrid simulation. Earthq. Eng. Struct. Dyn. 42, 1697–1715 (2013). doi:10.1002/eqe.2294
[11] Gunay, S., Mosalam, K.M.: Enhancement of real-time hybrid simulation on a shaking table configuration with implementation of an advanced control method. Earthq. Eng. Struct. Dyn. 44, 657–675 (2015). doi:10.1002/eqe.2477
[12] Schellenberg, A.H., Mahin, S.A., Fenves, G.L.: Advanced Implementation of Hybrid Simulation. (2009)
[13] McCrum, D.P., Williams, M.S.: An overview of seismic hybrid testing of engineering structures. Eng. Struct. 118, 240–261 (2016). doi:10.1016/j.engstruct.2016.03.039
[14] Géron, A.: Hands-on Machine Learning with Scikit-Learn and TensorFlow. (2017)
[15] Spiridonakos, M.D., Chatzi, E.N.: Metamodeling of dynamic nonlinear structural systems through polynomial chaos NARX models. Comput. Struct. 157, 99–113 (2015). doi:10.1016/j.compstruc.2015.05.002
[16] Chen, C., Ricles, J.M.: Servo-hydraulic actuator control for real-time hybrid simulation. 5222–5227 (2009). doi:10.1109/ACC.2009.5160186
[17] OpenSees. "Open System for Earthquake Engineering Simulation." from http://opensees.berkeley.edu., (2008)
[18] Bas, E.E., Moustafa, M.A., Pekcan, G. Compact Hybrid Simulation System: Validation and Applications for Braced Frame Seismic Testing. J. Earthq. Eng. (revised verison under review)
[19] Schellenberg, A.H., Kim, H.K., Mahin, S.A.: OpenFresco. Universtiy of California, Berkeley., (2009)
[20] MTS. "Civil engineering testing solutions." from http://www.mts.com. (2014)
[21] Sak, H., Senior, A., Beaufays, F.: Long Short-Term Memory Recurrent Neural Network Architectures for Large Scale Acoustic Modeling. Int. J. Speech Technol. 22, 21–30 (2019). doi:10.1007/s10772-018-09573-7
[22] Chang, S.-Y.: An Explicit Method with Improved Stability Property. Proc. 2011 Am. Control Conf. 1885–1891 (2009). doi:10.1002/nme